\PassOptionsToPackage{unicode}{hyperref}
\PassOptionsToPackage{hyphens}{url}
\documentclass[
]{article}
\usepackage{xcolor}
\usepackage{amsmath,amssymb}
\usepackage{iftex}
\ifPDFTeX
  \usepackage[T1]{fontenc}
  \usepackage[utf8]{inputenc}
  \usepackage{textcomp} % provide euro and other symbols
\else % if luatex or xetex
  \usepackage{unicode-math} % this also loads fontspec
  \defaultfontfeatures{Scale=MatchLowercase}
  \defaultfontfeatures[\rmfamily]{Ligatures=TeX,Scale=1}
\fi
\usepackage{lmodern}
\ifPDFTeX\else
\fi
\IfFileExists{upquote.sty}{\usepackage{upquote}}{}
\IfFileExists{microtype.sty}{% use microtype if available
  \usepackage[]{microtype}
  \UseMicrotypeSet[protrusion]{basicmath} % disable protrusion for tt fonts
}{}
\makeatletter
\@ifundefined{KOMAClassName}{% if non-KOMA class
  \IfFileExists{parskip.sty}{%
    \usepackage{parskip}
  }{% else
    \setlength{\parindent}{0pt}
    \setlength{\parskip}{6pt plus 2pt minus 1pt}}
}{% if KOMA class
  \KOMAoptions{parskip=half}}
\makeatother
\usepackage{graphicx}
\makeatletter
\newsavebox\pandoc@box
\newcommand*\pandocbounded[1]{% scales image to fit in text height/width
  \sbox\pandoc@box{#1}%
  \Gscale@div\@tempa{\textheight}{\dimexpr\ht\pandoc@box+\dp\pandoc@box\relax}%
  \Gscale@div\@tempb{\linewidth}{\wd\pandoc@box}%
  \ifdim\@tempb\p@<\@tempa\p@\let\@tempa\@tempb\fi% select the smaller of both
  \ifdim\@tempa\p@<\p@\scalebox{\@tempa}{\usebox\pandoc@box}%
  \else\usebox{\pandoc@box}%
  \fi%
}
\def\fps@figure{htbp}
\makeatother
\usepackage[]{natbib}
\usepackage[letterpaper,margin=1in]{geometry}
\ifPDFTeX\usepackage{mathptmx}\fi
\makeatletter\@ifundefined{c@none}{}{}\makeatother
\usepackage{longtable,booktabs,array,calc}
\providecommand{\real}[1]{#1}
\renewcommand{\_}{\textunderscore\allowbreak}
\usepackage{newunicodechar}
\newunicodechar{–}{\textendash}
\newunicodechar{—}{\textemdash}
\newunicodechar{≈}{\ensuremath{\approx}}
\newunicodechar{⇒}{\ensuremath{\Rightarrow}}
\newunicodechar{→}{\ensuremath{\rightarrow}}
\newunicodechar{×}{\ensuremath{\times}}
\newunicodechar{≤}{\ensuremath{\leq}}
\newunicodechar{≥}{\ensuremath{\geq}}
\newunicodechar{∼}{\ensuremath{\sim}}
\newunicodechar{≪}{\ensuremath{\ll}}
\usepackage{bookmark}
\IfFileExists{xurl.sty}{\usepackage{xurl}}{} % add URL line breaks if available
\hypersetup{
  pdftitle={Conformal Orbit-Valid Trust Horizons for Equivariant World Models},
  pdfauthor={Hongbo Wang},
  hidelinks,
  pdfcreator={LaTeX via pandoc}}

\title{Conformal Orbit-Valid Trust Horizons for Equivariant World
Models}
\author{Hongbo Wang \\
  \small Department of Mathematics, Stony Brook University, Stony Brook, NY 11794, USA}
\date{}

\begin{document}
\maketitle
\begin{abstract}
Learned world models are useful only over horizons on which their
rollout error remains controlled. We study trust-horizon certification
for latent world models with known group symmetries. Given a one-step
latent residual and a finite-time expansion estimate, we form a raw
horizon curve and calibrate it with a split-conformal multiplicative
factor. On the reproducible audit set, the conformal factor is
\(\gamma_\alpha=1.0\): the raw certificate is already conservative under
the audit protocol. Across 50 stable audits, we observe zero
anti-conservative violations, corresponding to an exact-binomial 95\%
upper bound of 5.8\% on the violation rate. Our main structural result
is that exact equivariance transports a calibrated trust-horizon curve
over the group orbit: when the environment dynamics, encoder, predictor,
action transform, and latent metric satisfy the stated
equivariance/invariance conditions, rollout errors and trust horizons
are orbit-constant. Empirically, the implemented models exhibit small
orbit-transport residuals, with median 1.1\% and maximum 4.1\% over 14
orbit audits. The certificate is also non-vacuous (median
certified-to-measured horizon ratio 0.67). A certificate-level
calibration-cost study shows two complementary regimes. On a symmetric
2D substrate, equivariant, plain, and augmented models are all
orbit-valid from a single calibration sector --- no separation, because
the substrate already makes non-equivariant baselines approximately
orbit-robust. A 3D yaw audit shows the other regime: the equivariant
model obtains a one-sector safe and non-vacuous orbit-valid certificate,
while healthy non-equivariant baselines pay violation, slack, sharpness,
or additional-sector cost. The certificate is a conservative,
distributional audit rather than a global reachability guarantee, and
certificate-guided subgoal spacing is not confirmed in the current 3D
CEM-MPC behavior layer.
\end{abstract}

\subsection{1. Introduction}\label{introduction}

\textbf{1.1 Calibration is geometric.} Calibration is usually treated as
a statistical operation over sampled points. For equivariant world
models, calibration also has \emph{geometry}: once a trust-horizon
certificate is valid on a representative region, exact equivariance can
transport that certificate over the group orbit. This structural
statement does \textbf{not} imply that equivariance always yields a
large empirical calibration-cost advantage: on our symmetric 2D
substrate, non-equivariant baselines are already approximately
orbit-robust, so the realized separation is mild (Section~5).

\textbf{1.2 The trust-horizon problem.} Model-based planning,
imagination, and self-supervision all roll a learned dynamics model
forward --- but every learned model degrades, and \emph{when} it stops
being trustworthy is rarely answered in a principled way (FF-JEPA's
\(H{=}25\) \citep{ffjepa2026} is one number for one distribution). The
field splits ``generate pixels'' (expensive) from ``predict latents''
(JEPA); we work in the latent regime and ask: \textbf{can a latent world
model certify, a priori and with finite-sample semantics, how many steps
it can be trusted --- and does that certificate hold across a symmetry
group?}

\textbf{1.3 The approach.} A raw error-propagation curve gives a
\emph{candidate} trust horizon; we make it \emph{statistically
conservative} by split-conformal calibration (Proposition B) and
\emph{orbit-valid} by equivariance (Theorem A, Corollary C) --- a curve
calibrated on one wedge transports over the whole group orbit at the
\emph{same} confidence, because equivariance turns the orbit-wide
coverage event into a single calibrated event rather than a union over
transformations. A non-equivariant model can still be conformalized, but
earns coverage only where it was calibrated; what it lacks is
\emph{zero-shot orbit transport}. The advantage is thus not that
non-equivariant certificates are impossible, but that exact equivariance
\emph{removes the orbit-wise calibration cost} when the symmetry holds.

\textbf{1.4 What else we measure, and a boundary.} Beyond the
certificate we characterize what equivariance buys (a low-data
prediction edge that scale erases; a persistent group-coordinate
readability) and costs (high cross-seed variance), and find that on the
2D pixel substrate neither corpus size nor input resolution improves
latent predictive precision (Appendix D). The natural downstream use ---
certified subgoal spacing --- is \emph{not confirmed}: a from-scratch 3D
CEM-MPC planner is high-variance and at \(n{=}10\) shows no detectable
certificate-to-horizon association (underpowered; Section~6). We report
these boundaries rather than bury them.

\textbf{1.5 Contributions.} 1. \textbf{Conformal trust-horizon
certificate.} A raw latent error-propagation curve becomes a one-sided
conservative certificate via split-conformal calibration
(\(\gamma_\alpha{=}1.0\)); \(0/50\) anti-conservative audits,
exact-binomial \(95\%\) upper bound \(5.8\%\). 2. \textbf{Orbit-validity
theorem.} Proposition B (conformal coverage) + Theorem A (exact orbit
transport) \(\Rightarrow\) Corollary C: a wedge-calibrated certificate
is orbit-valid at the same \(1-\alpha\), without orbit-wise calibration.
3. \textbf{Sharpness and implementation residuals.} The certificate is
non-vacuous (median \(H_{\mathrm{conf}}/H_{\mathrm{meas}}=0.67\),
\(0/172\) checks) and the realized transport residual is small (median
\(1.1\%\), max \(4.1\%\), \(n{=}14\)). 4. \textbf{Substrate-dependent
calibration cost.} On a symmetric 2D substrate, equivariant, plain, and
augmented models are all orbit-valid from one sector --- no separation,
because the substrate already induces approximate orbit robustness. On a
3D yaw audit the equivariant model obtains a one-sector safe and
non-vacuous certificate, while healthy plain baselines pay violation,
slack, sharpness, or additional-sector cost --- so equivariance reduces
orbit-wise calibration cost where baselines lack that robustness.

\textbf{Scoping note.} The certificate machinery (Benettin
\(\hat\lambda_1\), dual-boundary gates) is shared with a companion audit
line that certifies \emph{off-the-shelf} world models; this paper's
distinct contributions are the \emph{conformal} calibration, the
\emph{orbit-transport theorem and its corollary}, the sharpness/residual
accounting, and the substrate-dependent calibration-cost boundary.

\subsection{2. Setup and Preliminaries}\label{setup-and-preliminaries}

\subsection{2.1 The equivariant world
model}\label{the-equivariant-world-model}

An observation \(x\in\mathcal X\) (pixels or a point cloud) is encoded
to a latent \(z=E(x)\in\mathcal Z\); an \textbf{action-conditioned
predictor} \(f:\mathcal Z\times\mathcal A\to\mathcal Z\) advances it,
\(\hat z_{t+1}=f(z_t,a_t)\); and at the coarse timescale a
\textbf{subgoal predictor} \(G\) produces an action-free latent flow
\(\hat z_{sg,m+1}=G(z_{sg,\le m})\) (FF-JEPA's object). The encoder is
trained self-supervised (EMA-target JEPA with a variance floor); the
deployable model is the \emph{(target-encoder, predictor)} pair.

A group \(g\in\mathcal G\) acts on observations; \(\rho(g)\) is the
induced (orthogonal) representation on latents and \(\sigma(g)\) the
action on \(\mathcal A\). The model is \textbf{equivariant} when

\[
E(g\cdot x)=\rho(g)\,E(x),\qquad f(\rho(g)z,\,\sigma(g)a)=\rho(g)\,f(z,a).
\]

We realize \(E\) as a \(C_N\)-steerable network (2D pixels) or a
Vector-Neuron / \(SE(3)\) network (3D point clouds); equivariance is
verified post-training by a unit test that applies a random \(g\) and
checks the residual
\(\lVert E(g\cdot x)-\rho(g)E(x)\rVert/\lVert E(x)\rVert\)
(machine-precision for \(C_N\), \(\sim10^{-4}\) for the learned 3D
case). The transports \(\rho(g)\) on the structured latent --- rotating
the vector block, fixing the invariants --- are exact and are what the
orbit-transport claims (Section~3.3) consume.

\subsection{2.2 The audit machinery}\label{the-audit-machinery}

From a held-out interaction set we measure two quantities of the
deployable pair. The \textbf{one-step bias}
\(\hat\delta=\mathrm{median}_t\lVert f(z_t,a_t)-z_{t+1}\rVert\) is the
latent prediction error the certificate integrates. The \textbf{leading
exponent} \(\hat\lambda_1\) is estimated by a Benettin/QR scheme on the
predictor Jacobian via forward-mode JVP through \(f\) only
(\texttt{window\_exponent}), giving the rate at which a rollout's error
grows. A \textbf{faithful dual-boundary gate} compares the certified
horizon to the measured one under one-sided (conservative) and two-sided
(calibration) semantics; the gate code is shared across the program and
equivalence-tested. All seeds are explicit; statistics are reported per
\emph{run} (the equivariant pipeline's cross-run variance, Section~4.3,
makes per-run accounting necessary), and weak effects use \(n\ge10\).

\subsection{2.3 The certified horizon and its two
regimes}\label{the-certified-horizon-and-its-two-regimes}

The certificate (developed in Section~3) is

\[
\widehat{\mathrm{Err}}(H)=\hat\delta\sum_{t=0}^{H-1}e^{\hat\lambda_1 t},\qquad
H^*(\epsilon)=\max\{H:\widehat{\mathrm{Err}}(H)\le\epsilon\}.
\]

Both regimes are in scope by construction. \textbf{Neutral}
(\(\hat\lambda_1\!\approx\!0\)): the sum degenerates to the linear
budget \(H\hat\delta\) and the certificate's job is to calibrate
\(\hat\delta\). \textbf{Expansive} (\(\hat\lambda_1\!>\!0\)): the
spectral term dominates and \(H^*\) shrinks like
\(\log(1/\epsilon)/\hat\lambda_1\). The quasi-static manipulation tasks
we study are predominantly neutral; we flag this because it is exactly
the regime where, as Section~6 finds, the certificate's \emph{value} for
planning is weakest --- a scoping fact, not a defect.

\subsection{2.4 Datasets and tasks}\label{datasets-and-tasks}

Encoders are trained on \(\kappa\)-uniform weak-policy interaction
corpora; 2D experiments use PushT (pixels) with a dynamics knob
\(\kappa\) (damping) controlling the neutral/expansive regime, 3D
experiments use ManiSkill point-cloud manipulation (PickCube, PushCube).
Fractions are episode-level over the encoder-training corpus; the
certified curve is evaluated against true multi-step measured rollouts
on held-out episodes.

\subsection{3. Conformal Orbit-Valid Trust
Horizons}\label{conformal-orbit-valid-trust-horizons}

\begin{figure}
\centering
\includegraphics[width=0.98\linewidth,height=\textheight,keepaspectratio]{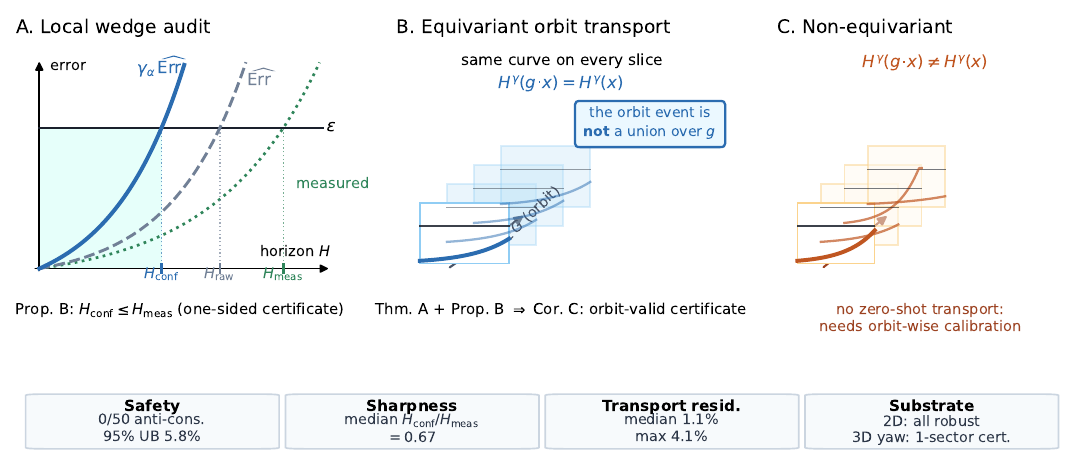}
\caption{\textbf{Certificate Cube.} A local split-conformal
trust-horizon curve is calibrated on one wedge (\(\gamma_\alpha\)
multiplies the whole error curve, so
\(H_{\mathrm{conf}}\le H_{\mathrm{meas}}\); Prop.~B).
Exact equivariance transports the \emph{same} calibrated curve over the
group orbit, so the orbit-wide coverage event is not a union over \(g\)
but one calibrated event viewed through the group action
(Thm.~A, Cor.~C). Without equivariance,
local calibration stays local and orbit-wise coverage must be paid for
separately. The certificate is a one-sided distributional audit, not a
global reachability guarantee.}
\end{figure}

\subsection{3.1 Raw curve and the conformal one-sided
certificate}\label{raw-curve-and-the-conformal-one-sided-certificate}

Let \(E:\mathcal X\to\mathcal Z\) be the encoder and
\(f:\mathcal Z\times\mathcal A\to\mathcal Z\) the action-conditioned
predictor. From held-out audits we measure a one-step residual
\(\hat\delta_i=\mathrm{median}\,\lVert f(z_t,a_t)-z_{t+1}\rVert\) and a
finite-time expansion estimate \(\hat\lambda_i\) (Benettin/QR on the
predictor Jacobian), giving the \textbf{raw horizon curve}
\(\widehat{\mathrm{Err}}_i(H)=\hat\delta_i\sum_{t<H}e^{\hat\lambda_i t}\)
and
\(H^{\mathrm{raw}}_i(\epsilon)=\max\{H:\widehat{\mathrm{Err}}_i\le\epsilon\}\).
This is an \emph{estimator}, not a guarantee:
\(\hat\delta_i,\hat\lambda_i\) are empirical, not uniform upper bounds.
We calibrate its one-sided conservatism with a multiplicative
split-conformal factor on the whole curve,
\(\widehat{\mathrm{Err}}^{\mathrm{conf}}_i=\gamma_\alpha\widehat{\mathrm{Err}}_i\),
\(H^{\mathrm{conf}}_i=\max\{H:\widehat{\mathrm{Err}}^{\mathrm{conf}}_i\le\epsilon\}\).

\textbf{Proposition B (one-sided conformal coverage).} \emph{Let
\(H^{\gamma}_i(\epsilon)=\max\{H:\gamma\,\widehat{\mathrm{Err}}_i(H)\le\epsilon\}\)
be the horizon under a curve multiplier \(\gamma\ge 1\) (non-increasing
in \(\gamma\), with \(H^{1}_i=H^{\mathrm{raw}}_i\) and
\(H^{\gamma_\alpha}_i=H^{\mathrm{conf}}_i\)), and take as nonconformity
score the minimal multiplier that makes the certified horizon
conservative,
\(\Gamma_i(\epsilon)=\inf\{\gamma\ge 1:H^{\gamma}_i(\epsilon)\le H^{\mathrm{meas}}_i(\epsilon)\}\).
With \(\gamma_\alpha=Q_{1-\alpha}(\Gamma_{1:n})\) on a calibration
split, a fresh exchangeable audit satisfies
\(\Pr(H^{\mathrm{conf}}(\epsilon)\le H^{\mathrm{meas}}(\epsilon))\ge 1-\alpha\).}
(Proof: monotonicity in \(\gamma\) gives
\(H^{\mathrm{conf}}\le H^{\mathrm{meas}}\iff\gamma_\alpha\ge\Gamma\),
then split-conformal quantile coverage. Appendix A.2.)

On the reproducible audit set every \(\Gamma_i=1\) --- the raw curve
already meets the one-sided condition
\(H^{\mathrm{raw}}\le H^{\mathrm{meas}}\) under the audit protocol ---
so \(\gamma_\alpha=1.0\) and calibration supplies the
\emph{finite-sample semantics}, not extra slack --- it is what prevents
the raw median-\(\hat\delta\) curve from being presented as an
uncalibrated guarantee. The guarantee is relative to the measured audit
distribution, \textbf{not} a global worst-case reachability bound.

\subsection{3.2 Exact orbit transport}\label{exact-orbit-transport}

Equivariance contributes a separate, deterministic property --- it
carries the calibrated curve over the whole group orbit for free.

\textbf{Theorem A (orbit transport).} \emph{Let \(G\) act by
\(x\mapsto g\!\cdot\!x\), \(a\mapsto\sigma(g)a\), \(z\mapsto\rho(g)z\)
with \(\rho(g)\) norm-preserving, and let \(E,T,f\) be equivariant
(\(E(g\!\cdot\!x)=\rho(g)E(x)\),
\(T(\rho(g)z,\sigma(g)a)=\rho(g)T(z,a)\),
\(f(\rho(g)z,\sigma(g)a)=\rho(g)f(z,a)\)). Then rollout errors are
orbit-invariant,
\(\lVert\hat z_t^{\,g}-z_t^{\,g}\rVert=\lVert\hat z_t-z_t\rVert\) for
all \(t\), so \(\hat\delta\), \(\hat\lambda\), the whole curve
\(\widehat{\mathrm{Err}}\), and hence \(H^{\mathrm{meas}}\) together
with \(H^{\gamma}\) for every fixed multiplier \(\gamma\ge 1\) --- in
particular the raw horizon \(H^{\mathrm{raw}}\) and the calibrated
\(H^{\mathrm{conf}}=H^{\gamma_\alpha}\) (\(\gamma_\alpha\) a single
calibration-set scalar) --- are constant along the orbit.} (Induction on
\(t\); full proof in Appendix B.)

\subsection{3.3 Orbit-valid conformal
certificate}\label{orbit-valid-conformal-certificate}

Combining the statistical and structural layers gives the paper's
central result.

\textbf{Corollary C (orbit-valid certificate).} \emph{If the certificate
is calibrated (Prop. B) on audits from a fundamental region and exact
equivariance (Thm. A) holds, then for a fresh audit \(x\) in that
region} \[
\Pr\!\bigl(\,H^{\mathrm{conf}}(g\!\cdot\!x,\sigma(g)a)\le H^{\mathrm{meas}}(g\!\cdot\!x,\sigma(g)a)
\ \ \forall g\in G\,\bigr)\ \ge\ 1-\alpha .
\]

The proof is the message: by Theorem A both horizons are orbit-constant,
so the orbit-wide event is \textbf{not a union over \(g\)} --- it is the
single pointwise event
\(\{H^{\mathrm{conf}}(x)\le H^{\mathrm{meas}}(x)\}\) transported along
the orbit. Hence \emph{wedge-level} coverage \(1-\alpha\) \textbf{is}
\emph{orbit-level} coverage \(1-\alpha\): equivariance removes the
orbit-wise calibration cost --- no union bound, no orbit-spread
calibration data --- when the symmetry holds. A non-equivariant model
can still be conformalized, but earns coverage only where it was
calibrated; off that wedge it must collect fresh exchangeable data or
assume extra smoothness.

\textbf{Which group acts on the dynamics.} We distinguish
representation-level equivariance from dynamics-level transport. On a 3D
tabletop, gravity and contact break full \(SO(3)\), so Corollary C
applies at the dynamics level only on the symmetry subgroup that
preserves the transition \emph{and} action semantics (e.g.~yaw
\(SO(2)\), or controlled synthetic point-cloud rotations) --- not
arbitrary physical \(SO(3)\).

\subsection{3.4 Approximate implementation
residual}\label{approximate-implementation-residual}

Trained networks are only approximately equivariant, so transport is
approximate.

\textbf{Proposition D (sketch).} \emph{With encoder/predictor
equivariance defects \(\eta_E,\eta_f\) and an \(L\)-Lipschitz predictor,
\(\bigl|\,e_t^{\,g}-e_t\,\bigr|\le R_t=L^{t}\eta_E+\eta_f\sum_{k<t}L^{k}+b_t\),
which for the neutral regime \(L\approx 1\) reduces to
\(R_t\lesssim\eta_E+t\,\eta_f+b_t\) (target residual \(b_t=0\) exact,
\(\le\eta_E\) under bounded encoder residual). The inflated orbit
certificate \(\gamma_\alpha\widehat{\mathrm{Err}}+R_H\) restores
one-sided coverage over the orbit.} (Full statement + proof in Appendix
B.)

Empirically the realized transport residual is small --- median
\(1.1\%\), max \(4.1\%\) over \(14\) orbit audits (the quantity Prop. D
bounds; a separate, looser out-of-wedge degradation diagnostic,
\emph{without} transport, is median \(2.2\%\) / max \(6.0\%\)). Since
\(\eta_E\sim10^{-4}\) (the Section~2.1 equivariance unit test) and the
studied tasks are neutral (\(L\approx 1\)), \(R_t\) stays small over
\(H\le 8\).

\subsection{4. Empirical Audit and
Sharpness}\label{empirical-audit-and-sharpness}

We audit the conformal certificate on a \textbf{reproducible audit set}
--- audits passing pre-specified stability/reproducibility filters
applied \emph{before} the anti-conservatism indicator is evaluated, so
filtering cannot select for or against violations (the \(75\!\to\!50\)
filter-stage accounting is in Appendix A).

\textbf{Safety (audit-level).} Over the \(50\) stable audit cells we
observe \(0/50\) anti-conservative violations at
\(\gamma_\alpha{=}1.0\); the exact-binomial \(95\%\) upper bound for a
zero-failure sample of size \(50\) is \(1-0.05^{1/50}=5.8\%\).
Per-family bounds are wide for small \(n\) (the \(n{=}8\) subset alone
gives a \(31.2\%\) upper bound), so the pooled set carries the strength;
the breakdown is in Appendix A. As a \emph{supporting} held-out
diagnostic, the calibrated curve passes \(86/86\) one-sided coverage
checks at the finer curve level --- reported separately because
curve-level checks within an audit are not independent and would inflate
the effective sample size if treated as the primary denominator.

\textbf{Sharpness (is it safe because useless?).} No.~As a point-level
diagnostic over \(172\) cell\(\times\)tolerance checks, the
certified-to-measured horizon ratio
\(H^{\mathrm{conf}}/H^{\mathrm{meas}}\) has \textbf{median \(0.67\), IQR
\([0.38,1.0]\)}, with \(0/172\) anti-conservative; \(98\%\) of checks
retain \(\ge 25\%\) of the measured horizon, \(69\%\) retain
\(\ge 50\%\), and \(27\%\) are \emph{exactly tight}
(\(H^{\mathrm{conf}}{=}H^{\mathrm{meas}}\), including correctly
certifying the full \(8\)-step horizon). The certificate is one-sided
conservative \emph{and} non-vacuous. Horizons are short integers
(\(H\in[1,8]\), the neutral regime), so it is best read as a sharp
short-horizon feasibility gate rather than a long-horizon point
predictor.

\begin{figure}
\centering
\includegraphics[width=0.86\linewidth,height=\textheight,keepaspectratio]{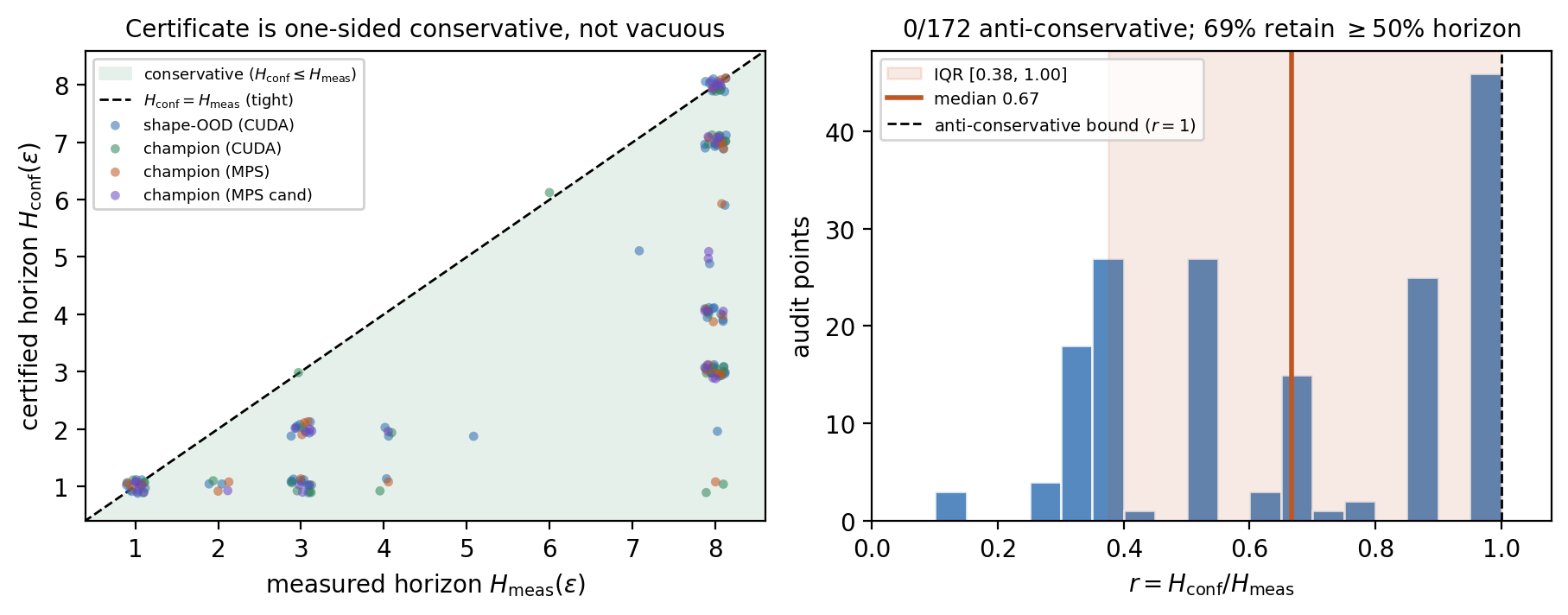}
\caption{The certificate is one-sided conservative and non-vacuous.
Left: certified vs.~measured horizon over \(172\) audit checks --- every
point in the conservative region, with real spread (at
\(H_{\mathrm{meas}}{=}8\) the certified horizon ranges \(1\)--\(8\),
including tight points). Right: the ratio
\(H_{\mathrm{conf}}/H_{\mathrm{meas}}\) (median \(0.67\), IQR
\([0.38,1.0]\), \(0/172\) anti-conservative). Point-level diagnostic;
the formal audit-level safety accounting remains the \(50\)-cell table.}
\end{figure}

\subsection{5. Calibration Cost Across
Substrates}\label{calibration-cost-across-substrates}

Corollary C predicts that equivariance removes the orbit-wise
calibration cost. We test this \emph{directly} at the certificate level:
place a held-out set at each of \(K\) orbit sectors (the group action on
inputs and actions), audit each sector, then calibrate \(\gamma\) on
\(m\) sectors and ask whether the certificate stays one-sided valid over
the \textbf{whole} orbit (calibration and test sectors disjoint). This
is a \textbf{boundary test} --- it asks \emph{when} the predicted
structural saving becomes an empirical one --- and the answer depends on
the substrate.

\subsection{5.1 Symmetric 2D: no
separation}\label{symmetric-2d-no-separation}

On the circular-masked 2D PushT substrate, equivariant, plain
(non-equivariant), and R-augmented models \textbf{all} show \(0\%\)
orbit anti-conservative violations from a single calibration sector
(\(m{=}1\)); the orbit non-constancy of \(\hat\delta\) is small for all
three (spread \(\le 8\%\)), and the augmented model is in fact the
\emph{most} orbit-robust (Fig.~3A). The theorem-side
transport holds (Section~3.4), but this substrate does \textbf{not}
expose a calibration-cost separation: a globally rotation-symmetric
arena makes even non-equivariant CNNs approximately orbit-robust for
free.

\subsection{5.2 3D yaw: the separation
appears}\label{d-yaw-the-separation-appears}

On a 3D point-cloud manipulation substrate audited under the
gravity-preserving yaw subgroup (\(SO(2)\); full \(SO(3)\) is a
representation-level symmetry but not a tabletop \emph{dynamics}
symmetry, Section~3.3), the plain baseline no longer receives orbit
robustness for free. All multipliers and violation rates in this
subsection use the \textbf{curve-level} score \(\Gamma_i\) of
Proposition B (the deprecated horizon-division proxy is retained only as
a sanity check; Appendix A.3). The equivariant Vector-Neuron model
obtains a \textbf{one-sector safe and non-vacuous} certificate,
consistently across \(3\) seeds: \(0\) anti-conservative violations at
every \(m\), \(\gamma_\alpha{=}1.0\), median
\(H^{\mathrm{conf}}/H^{\mathrm{meas}}=0.50\), utility-preserving
coverage \(75\%\) at \(\tau{=}0.4\), and minimal calibration sectors
\(m^*_{0.4}=1\) (Fig.~3B).

Healthy plain seeds do \textbf{not} obtain this certificate, and the
cost takes different forms across seeds. Plain seed r0 does not reach
orbit safety \emph{within the audited sector budget} \(m\in\{1,2,4\}\):
its violation rate falls from \(40\%\) to \(21\%\) to \(11\%\) but stays
anti-conservative, even though it already pays a slack multiplier
\(\gamma_\alpha\approx1.9\). The second healthy plain seed (r2) is
conservative (\(0\) violations) but loses \emph{sharpness} --- median
\(H^{\mathrm{conf}}/H^{\mathrm{meas}}\approx0.38\), below the
\(\tau{=}0.4\) utility threshold. A third plain seed fails the
pre-registered stability filter (latent collapse) and is reported
separately (Appendix A). So the non-equivariant baseline does not obtain
the one-sector safe-and-sharp certificate; the cost surfaces as
violation, slack, sharpness collapse, or seed instability.

These residual plain violations are \emph{expected}, not a breakdown of
the conformal machinery: for a non-equivariant model the locally
calibrated sector is not guaranteed exchangeable with the rest of the
orbit, so a wedge-calibrated certificate carries no zero-shot transport
(Corollary C needs the equivariance hypothesis). The remaining
violations are precisely the empirical price of that missing structure.
The 3D audit otherwise uses the same certificate definition as the rest
of the paper: the GPU rollout implementation is verified equal to the
canonical CPU audit on shared cells (\(\hat\delta\) \(4.6526\) vs
\(4.653\); identical horizon cells; Appendix A).

\subsection{5.3 Takeaway}\label{takeaway}

The two substrates are complementary regimes. The orbit-transport
theorem is \emph{structural} and always holds; the \emph{empirical}
calibration-cost saving appears only when ordinary baselines do not
already acquire orbit robustness from the data distribution. On the
symmetric 2D arena they do, so the separation is null; on the 3D yaw
substrate they do not, so equivariance yields a one-sector certificate
that non-equivariant baselines cannot stably match. \emph{Calibration
has geometry, but whether that geometry pays off empirically is a
property of the substrate.}

\begin{figure}
\centering
\includegraphics[width=0.98\linewidth,height=\textheight,keepaspectratio]{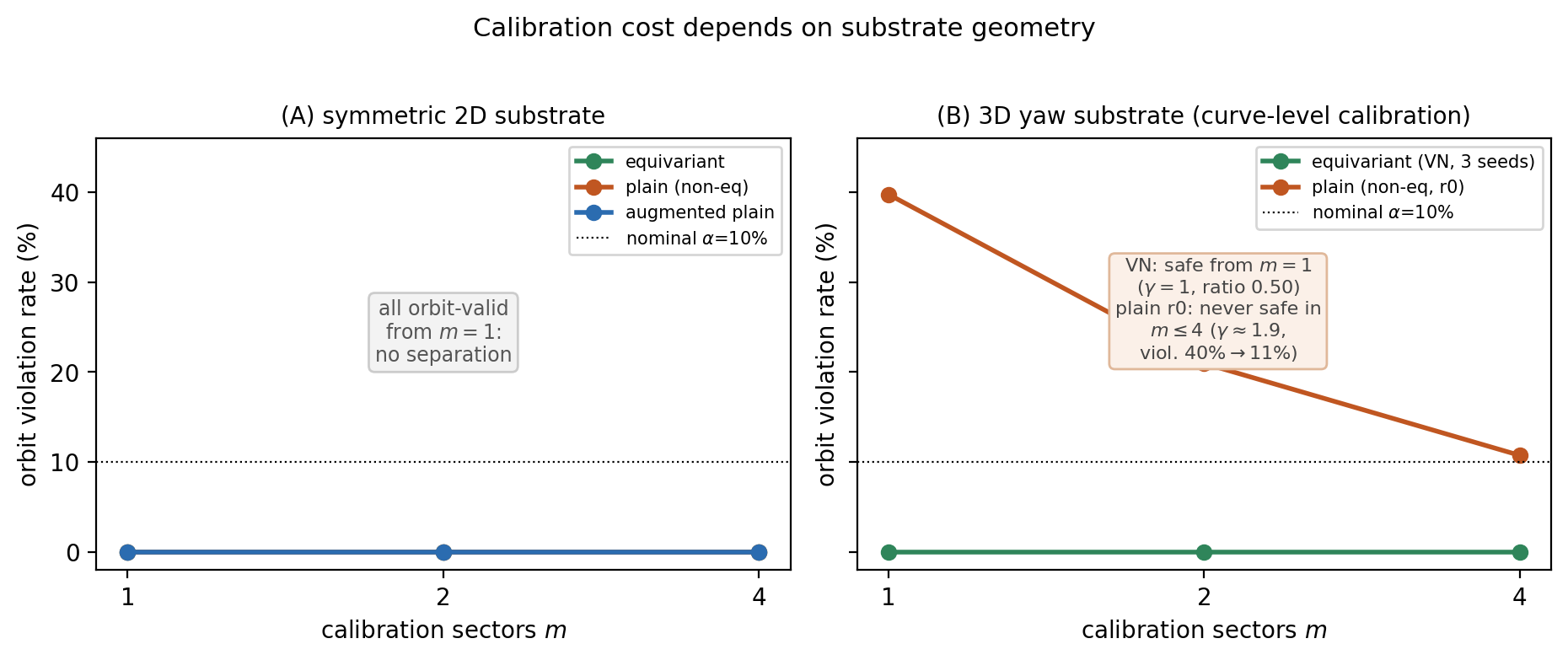}
\caption{Calibration cost depends on substrate geometry (both panels are
audit-time evaluations of frozen checkpoints --- no retraining;
violation rates use the curve-level multiplier \(H^{\gamma}\) of
Proposition B, not the deprecated horizon-division proxy). \textbf{(A)}
Symmetric 2D substrate: equivariant, plain, and augmented models are all
orbit-valid from one calibration sector --- no separation, because the
circular-masked arena already makes non-equivariant baselines
approximately orbit-robust. \textbf{(B)} 3D yaw substrate
(gravity-preserving \(SO(2)\) subgroup, not full \(SO(3)\)): the
equivariant VN family is safe from one sector (\(\gamma_\alpha{=}1\),
median \(H^{\mathrm{conf}}/H^{\mathrm{meas}}=0.50\), \(0\) violations at
every \(m\)), while the non-equivariant plain seed r0 stays
anti-conservative across the audited budget \(m\in\{1,2,4\}\)
(\(40\%\!\to\!21\%\!\to\!11\%\), slack \(\gamma_\alpha\approx1.9\),
never below the nominal \(\alpha\)). A second plain seed (r2) is
conservative but loses sharpness (median ratio \(0.38<\tau\)); a third
collapses (Appendix A).}
\end{figure}

\subsection{6. Scope and Boundaries}\label{scope-and-boundaries}

\textbf{What else the structure buys and costs.} Beyond the certificate,
equivariance shows a \emph{low-data} prediction edge that scale erases,
a \emph{persistent} readability/group-coordinate advantage, and a
\emph{high-variance} cost; on our 2D pixel substrate neither corpus size
nor input resolution improved latent predictive precision. We
characterize these in Appendix D. The binding finding there ---
equivariant world models are high-variance estimators --- is the cost
side that the planning non-confirmation below makes concrete.

The certified horizon \(H^*(\epsilon)\) is, by construction, a planning
quantity: it states how far a hierarchical controller may space its
subgoals before the model's error exceeds tolerance. The natural
application --- \emph{certified subgoal spacing}, replacing FF-JEPA's
fixed \(H{=}25\) with a data-derived, orbit-valid function --- is also
where structure must pay off as \emph{behavior}. We report this attempt
in full, including its negative result, because the boundary is itself
the finding.

\subsection{6.1 The substrate test}\label{the-substrate-test}

We built a from-scratch latent CEM-MPC planner that rolls the
equivariant predictor \(f\) forward over a horizon \(H\), scores action
sequences by the predicted read-out distance to the goal, and re-plans
(3D point-cloud PickCube, reusing the paper's machinery). The
pre-registered, certificate-aware question was minimal: \emph{does
latent CEM-MPC beat open-loop/random, and does \(H^*(\epsilon)\) predict
the optimal \(H\)?}

\subsection{6.2 A single-seed signal that did not survive
multiplicity}\label{a-single-seed-signal-that-did-not-survive-multiplicity}

On the first seed the result was clean and certificate-consistent: blind
planning at \(H{=}8\) drifted \emph{away} from the goal, while gating
the horizon to \(H{\approx}3\text{–}4\) --- exactly where the
predictor's measured rollout error crosses the \(\epsilon{\approx}0.1\)
tolerance (\(H^*(\epsilon)\)) --- produced a positive, peaked
\emph{reduction-vs-\(H\)} curve that fell off as the horizon entered the
untrustworthy region. Taken alone, this is an initial positive signal.

\textbf{It did not replicate.} Under the project's discipline
(\(n\ge10\) for weak effects), a well-powered sweep (\(n{=}10\) encoders
at \(400\) episodes) shows the planning substrate is
\textbf{high-variance}: \(3/10\) seeds fail at every horizon and the
per-seed optimum spreads across \(H\in\{2,3,4,8\}\). A formal test finds
no detectable trust-horizon / best-horizon association
(\(\tau_b{=}-0.15\)), but it is underpowered, so this is \emph{no
evidence of a link}, not a demonstrated absence (Appendix C). The
\textbf{binding, well-powered finding is the variance}: more data tames
the bias but leaves the cross-seed std in \(0.06\)--\(0.10\) across
\(150\to1000\) episodes (Section~4.3) --- the single-seed signal was a
seed-dependent effect. (A contact-rich variant, PushCube, behaved no
differently, with \(7\text{–}13\%\) first non-zero success.) This is the
\emph{cost} side of the structure--variance economics (Section~4) made
concrete.

\textbf{Scoping the negative.} This is \emph{not} a claim that latent
planning is infeasible: concurrent hierarchical latent planners reach
\(\sim70\%\) on real-robot pick-and-place \citep{hwm2026}. Our narrower
negative: \emph{certificate-gated, single-level} CEM-MPC on a
\emph{high-variance equivariant} substrate does not deliver. Whether
certified spacing can be made to pay off remains open.

\subsection{6.3 Diagnosis: a state-sufficiency
hypothesis}\label{diagnosis-a-state-sufficiency-hypothesis}

The failure pattern has a precise structural reading. Our predictor's
\emph{action-response} --- how accurately \(f(z,a)\) moves the read-out
end-effector in the direction the true end-effector moves --- is the
weak, high-variance component (cosine \(0.30\text{–}0.60\) across
seeds), while the encoder's \emph{static} quality (latent norm,
coordinate readability) is stable. A planner needs to predict
\emph{dynamics}; our latent carries only \emph{configuration}. A direct
symptom: an action-mismatch broken-equivariance control is uninformative
here, because zeroing the action leaves the one-step residual unchanged
(\(\hat\delta_{\text{zero-act}}=\hat\delta_{\text{real-act}}\)) --- the
predictor barely uses the action. (This is why Section~5 uses the
non-equivariant checkpoint, not an action mismatch, as its
representation-broken control.)

This is \textbf{consistent with a state-sufficiency bottleneck} (a
hypothesis motivated by the failure mode, not a demonstrated cause): we
impose symmetry on the \emph{configuration} group \(SE(3)\), but a world
model is a \emph{dynamics} object, and a position-centric latent may
under-determine the rollout --- one cannot predict how an object moves
from \emph{where} it is without \emph{how fast} it is going. A sharp
asymmetry sharpens it: on a tabletop, gravity breaks the spatial
rotation the architecture leans on (\(SO(3)\!\to\!SO(2)\)) yet preserves
Galilean boost --- the symmetry we never exploit.

\textbf{Future work.} We therefore leave certified planning to a
phase-space-equivariant substrate (latent carrying velocity, predictor
Galilean-/boost-equivariant), with a clean falsification: \emph{does
adding the velocity state and boost equivariance reduce the planning
variance and recover the certificate\(\to\)horizon link?} --- stated as
a hypothesis, not a result.

\subsection{7. Related Work and
Positioning}\label{related-work-and-positioning}

\textbf{Latent world models and hierarchical planning.} We work in the
latent-prediction regime {[}JEPA; \citet{lecun2022}; \citet{vjepa2024};
\citet{dreamerv3}{]} rather than pixel generation, which we argue
(Appendix D) is the limiting abstraction, in our setup, for a
\emph{predictive} model. The closest neighbor is FF-JEPA
\citep{ffjepa2026}, hierarchical latent planning with an action-free
subgoal predictor: it establishes the \emph{planning-layer demand} but
leaves the subgoal gap \(H{=}25\) an unprincipled hyperparameter. Our
\(H^*(\epsilon)\) is the principled answer it lacks --- \emph{one number
for one distribution} becomes \emph{a function valid over the orbit} ---
which is also why our negative planning result (Section~6) does not
collide with FF-JEPA: we share the demand, not a working planner. HWM
\citep{hwm2026} supplies a \emph{post-hoc} error-vs-horizon analysis
(its Fig. 6) that is the empirical shadow of our certified curve; the
differentiator is \emph{a priori, orbit-transported derivation} vs
post-hoc measurement.

\textbf{Equivariance, geometric deep learning, and generalization.} The
encoder builds on \(E(n)\)/steerable and Vector-Neuron equivariance
\citep{cohenwelling2016, satorras2021, vectorneurons2021, e3nn2022}, and
equivariant RL and policies
\citep{equivrl2022, equivrl_po2024, equibot2024} share the
shared-encoder pattern we reuse. The generalization benefit of
equivariance is partly understood theoretically --- a provably strict
generalization gain on symmetric targets \citep{elesedy2021} --- and
empirically contested at scale \citep{brehmer2024}. That literature
measures the benefit on a \emph{prediction} loss; our claim is
orthogonal to both architecture design and accuracy: Theorem A
transports a \emph{calibrated certificate} over the orbit, so a local
audit becomes orbit-valid at no extra calibration cost --- a statement
about the certificate's \emph{domain of validity}, which
prediction-generalization results do not address. Consistent with the
project's red-team discipline we do \textbf{not} claim ``equivariance
wins'': Appendix D shows the prediction edge is low-data (the Brehmer
objection is \emph{correct} on that axis) and Section~6 reports the
planning negative in full.

\textbf{Conformal prediction.} Conformal prediction yields
distribution-free, finite-sample validity under exchangeability
\citep{vovk2005, angelopoulos2023}, with split-conformal and
conformalized quantile regression as workhorses \citep{romano2019cqr}
and adaptive variants that relax exchangeability under distribution
shift \citep{gibbs2021adaptive}; it has recently been applied to
world-model verification \citep{geng2026}. We apply a one-sided
multiplicative split-conformal calibrator to the \emph{entire}
trust-horizon curve (Section~3.2). What is new is the \emph{object}, not
the calibrator: prior work calibrates a scalar or an interval at a
point, whereas we calibrate a horizon \emph{function} and then transport
that calibrated function over the group orbit (Theorem A) --- a local
audit becomes an orbit-valid certificate. A non-equivariant predictor
can be conformalized too, but earns coverage only where it was
calibrated; off that wedge it must collect fresh exchangeable data --- a
cost equivariance removes when the symmetry holds.

\textbf{Reachability, barrier functions, and runtime assurance.} Formal
safety has a long control-theoretic tradition: Hamilton--Jacobi
reachability computes backward-reachable sets for known dynamics
\citep{bansal2017hj}, control barrier functions enforce
forward-invariant safe sets \citep{ames2019cbf}, and runtime-assurance
architectures switch to a verified fallback when a controller leaves its
trusted envelope \citep{sha2001simplex}. These certify a
\emph{controller} against a \emph{known or bounded} model; our object is
dual --- \emph{how long a learned model can be trusted} before rollout
error breaks tolerance, with finite-sample conformal semantics rather
than a worst-case reachable set. Such guarantees are stated per state or
per trajectory; equivariance lets us state ours once per orbit, whereas
a non-equivariant trust certificate must re-establish its smoothness
budget across the orbit. The discrete-symmetry and companion neighbors
are symmetry-protected neutral Lyapunov modes \citep{mo2026} and an
audit line \citep{auditline} that certifies off-the-shelf models at zero
training cost.

\subsection{Relation to the companion audit
line}\label{relation-to-the-companion-audit-line}

The certificate \emph{machinery} --- the Benettin \(\hat\lambda_1\)
expansion estimate and the dual-boundary feasibility gates --- is
\textbf{shared} with the companion audit line \citep{auditline} that
certifies off-the-shelf models; we do not claim it as novel. What is
distinct here is the \emph{conformal} one-sided calibration of the
horizon curve (Section~3.2), the \emph{orbit-transport theorem} for the
calibrated curve (Section~3.3), the measured \emph{structure--variance
economics} (Appendix D), and the honest \emph{downstream boundary}
(Section~6). We are explicit about what this is and is not: a
\emph{certified-horizons + honest-economics} paper, not a
\emph{certified-spacing-wins} paper (Section~8 states the defensible
result and its over-reach boundary).

\subsection{8. Conclusion}\label{conclusion}

We introduced \emph{conformal orbit-valid trust horizons} for
equivariant world models. The certificate is one-sided and
distributional: it is calibrated relative to the measured audit
distribution, not a global reachability guarantee. The key structural
result is that exact equivariance transports conformal validity over a
group orbit, so the orbit-wide event is not a union over transformations
but the same calibrated event viewed through the group action.
Empirically the certificate is safe and non-vacuous: \(0/50\)
audit-level anti-conservative violations, a \(95\%\) upper bound of
\(5.8\%\), and median sharpness
\(H_{\mathrm{conf}}/H_{\mathrm{meas}}=0.67\); implementation-level
transport residuals are small (median \(1.1\%\)). At the same time, the
symmetric 2D substrate is already orbit-robust for non-equivariant and
augmented baselines, so the calibration-cost separation is mild rather
than dramatic. The natural downstream use --- certified subgoal spacing
--- remains a separate control-layer problem, not confirmed in our
high-variance 3D planner. The main lesson is that equivariance does not
replace calibration; it gives calibration a geometry.

\bibliography{paper3}

\clearpage
\appendix
\subsection{A. Audit accounting and the reproducible
set}\label{a.-audit-accounting-and-the-reproducible-set}

We report results on the \textbf{reproducible audit set} rather than the
looser candidate count. The set is defined by pre-specified
stability/reproducibility filters applied \textbf{before} the
anti-conservatism indicator \(V_i\) is evaluated, so the filtering
cannot select for or against violations.

{\def\LTcaptype{none} % do not increment counter
\begin{longtable}[]{@{}
  >{\raggedright\arraybackslash}p{(\linewidth - 6\tabcolsep) * \real{0.2308}}
  >{\raggedleft\arraybackslash}p{(\linewidth - 6\tabcolsep) * \real{0.3077}}
  >{\raggedright\arraybackslash}p{(\linewidth - 6\tabcolsep) * \real{0.2308}}
  >{\raggedright\arraybackslash}p{(\linewidth - 6\tabcolsep) * \real{0.2308}}@{}}
\toprule\noalign{}
\begin{minipage}[b]{\linewidth}\raggedright
Stage
\end{minipage} & \begin{minipage}[b]{\linewidth}\raggedleft
Count
\end{minipage} & \begin{minipage}[b]{\linewidth}\raggedright
Rule
\end{minipage} & \begin{minipage}[b]{\linewidth}\raggedright
Dep. on \(V_i\)?
\end{minipage} \\
\midrule\noalign{}
\endhead
\bottomrule\noalign{}
\endlastfoot
Candidate records & 75 & all generated audit artifacts & No \\
Reproducible & 50 & parseable certificate JSON, valid \(c_3\) curve &
No \\
Stable cells & 50 & stability criteria (no divergence, latent-std in
range, logs present) & No \\
\(V_i\) evaluated & 50 &
\(V_i=\mathbf 1\{H^{\mathrm{conf}}_i>H^{\mathrm{meas}}_i\}\) &
\textbf{Yes (post-filter)} \\
\end{longtable}
}

The \(75\to50\) reduction is \emph{not} a removal of failed-outcome
cells; it removes cells whose audit object is a-priori invalid (failed
training, missing or unparseable artifacts). Filtered-out cases and
their reasons are listed in the artifact
\texttt{papers/tables/audit\_conformal.json}.

\subsubsection{A.1 Full per-family
table}\label{a.1-full-per-family-table}

{\def\LTcaptype{none} % do not increment counter
\begin{longtable}[]{@{}
  >{\raggedright\arraybackslash}p{(\linewidth - 14\tabcolsep) * \real{0.0968}}
  >{\raggedleft\arraybackslash}p{(\linewidth - 14\tabcolsep) * \real{0.1290}}
  >{\raggedleft\arraybackslash}p{(\linewidth - 14\tabcolsep) * \real{0.1290}}
  >{\raggedleft\arraybackslash}p{(\linewidth - 14\tabcolsep) * \real{0.1290}}
  >{\raggedleft\arraybackslash}p{(\linewidth - 14\tabcolsep) * \real{0.1290}}
  >{\raggedleft\arraybackslash}p{(\linewidth - 14\tabcolsep) * \real{0.1290}}
  >{\raggedleft\arraybackslash}p{(\linewidth - 14\tabcolsep) * \real{0.1290}}
  >{\raggedleft\arraybackslash}p{(\linewidth - 14\tabcolsep) * \real{0.1290}}@{}}
\toprule\noalign{}
\begin{minipage}[b]{\linewidth}\raggedright
Family
\end{minipage} & \begin{minipage}[b]{\linewidth}\raggedleft
\(n\)
\end{minipage} & \begin{minipage}[b]{\linewidth}\raggedleft
\(\gamma_\alpha\)
\end{minipage} & \begin{minipage}[b]{\linewidth}\raggedleft
viol.
\end{minipage} & \begin{minipage}[b]{\linewidth}\raggedleft
med \(H_c/H_m\)
\end{minipage} & \begin{minipage}[b]{\linewidth}\raggedleft
med \(\hat\delta\)
\end{minipage} & \begin{minipage}[b]{\linewidth}\raggedleft
med \(\hat\lambda_1\)
\end{minipage} & \begin{minipage}[b]{\linewidth}\raggedleft
95\% UB
\end{minipage} \\
\midrule\noalign{}
\endhead
\bottomrule\noalign{}
\endlastfoot
Shape-OOD (CUDA) & 23 & 1.0 & 0/23 & 0.67 & 2.65 & 0.185 & 12.2\% \\
Champ (CUDA) & 10 & 1.0 & 0/10 & 0.50 & 2.51 & 0.192 & 25.9\% \\
Champ (MPS) & 8 & 1.0 & 0/8 & 0.67 & 3.05 & 0.211 & 31.2\% \\
Champ (MPS-cand) & 9 & 1.0 & 0/9 & 0.67 & 2.49 & 0.212 & 28.3\% \\
\textbf{Total} & \textbf{50} & \textbf{1.0} & \textbf{0/50} & --- & ---
& --- & \textbf{5.8\%} \\
Orbit-transp. resid. & 14 & --- & --- & 1.1\% & --- & --- &
4.1\%\(^\ddagger\) \\
\end{longtable}
}

Families correspond to the recorded shape-OOD, champion (CUDA / MPS /
MPS-cand), and wedge audit JSONs (stems in
\texttt{audit\_conformal.py}). The small per-family bounds are
deliberately reported: with \(n{=}8\) the \(95\%\) upper bound is
\(31.2\%\), so any single small family is weak evidence; the strength is
the pooled \(n{=}50\) (\(5.8\%\)). Exact zero-failure binomial bound:
\(1-0.05^{1/n}\). (\(^\ddagger\) the orbit-transport row reports median
/ max \emph{transport residual}, not a binomial bound.)

\subsubsection{A.2 Conformal calibration
details}\label{a.2-conformal-calibration-details}

The split-conformal multiplier is selected on a calibration split of
audit \((\text{cell},\epsilon)\) pairs and evaluated on a disjoint test
split. Consistent with the multiplicative \emph{curve} calibration
\(\widehat{\mathrm{Err}}^{\mathrm{conf}}=\gamma_\alpha\widehat{\mathrm{Err}}\)
(Section~3.1), the nonconformity score is the minimal curve multiplier
that makes the certified horizon conservative,
\(\Gamma_i=\inf\{\gamma\ge 1: H^{\gamma}_i\le H^{\mathrm{meas}}_i\}\)
with
\(H^{\gamma}_i=\max\{H:\gamma\widehat{\mathrm{Err}}_i(H)\le\epsilon\}\),
and \(\gamma_\alpha=Q_{1-\alpha}(\Gamma_{\mathrm{cal}})\),
\(\alpha=0.1\). On the reproducible audit set every recorded
\((\text{cell},\epsilon)\) already satisfies
\(H^{\mathrm{raw}}_i\le H^{\mathrm{meas}}_i\) (the raw curve is
one-sided conservative --- equivalently the recorded horizon ratio
\(H^{\mathrm{raw}}_i/H^{\mathrm{meas}}_i\le 1\)), so \(\Gamma_i=1\)
throughout and \(\gamma_\alpha=1.0\): no extra slack is required under
the audit protocol. (Where a violation occurs, \(\Gamma_i>1\) and the
curve multiplier differs from the bare horizon ratio; the two coincide
only in the all-conservative regime we record here.) This does not make
calibration vacuous --- it supplies the finite-sample coverage
\emph{interpretation} under exchangeability and prevents the raw
median-\(\hat\delta\) curve from being presented as an uncalibrated
guarantee.

\textbf{Denominator note.} Our \emph{primary} statistical accounting is
at the \textbf{run/audit level}: \(0/50\) anti-conservative audits,
\(95\%\) upper bound \(5.8\%\). At the finer \textbf{curve/checkpoint
level} the calibrated audit passes all \(86/86\) recorded one-sided
coverage checks on the held-out split; we report this only as supporting
diagnostic evidence, because curve-level checks within an audit are not
independent and would inflate the effective sample size if treated as
the primary denominator. Reproduce both with
\texttt{papers/figures/audit\_conformal.py}.

\subsubsection{A.3 3D yaw audit: health filter, per-seed cost, and
GPU/CPU
equivalence}\label{a.3-3d-yaw-audit-health-filter-per-seed-cost-and-gpucpu-equivalence}

The Section~5 3D result audits frozen checkpoints under the
gravity-preserving yaw subgroup \(SO(2)\) (no retrain, no simulation).
It is run on a separate point-cloud substrate (Vector-Neuron DGCNN
encoder for the equivariant family, PointNet for plain), so it inherits
its own stability filter rather than the 2D accounting above. We trained
three seeds per family and applied the \textbf{same} pre-registered
collapse filter used throughout the project (latent norm in range,
\(\hat\delta\) not degenerate) \emph{before} auditing.

{\def\LTcaptype{none} % do not increment counter
\begin{longtable}[]{@{}
  >{\raggedright\arraybackslash}p{(\linewidth - 4\tabcolsep) * \real{0.3333}}
  >{\raggedright\arraybackslash}p{(\linewidth - 4\tabcolsep) * \real{0.3333}}
  >{\raggedright\arraybackslash}p{(\linewidth - 4\tabcolsep) * \real{0.3333}}@{}}
\toprule\noalign{}
\begin{minipage}[b]{\linewidth}\raggedright
seed
\end{minipage} & \begin{minipage}[b]{\linewidth}\raggedright
VN (latent-norm / \(\hat\delta\))
\end{minipage} & \begin{minipage}[b]{\linewidth}\raggedright
plain (latent-norm / \(\hat\delta\))
\end{minipage} \\
\midrule\noalign{}
\endhead
\bottomrule\noalign{}
\endlastfoot
r0 & \(20.65\) / \(4.63\) --- healthy & \(38.60\) / \(5.92\) ---
healthy \\
r1 & \(19.50\) / \(5.06\) --- healthy & \(2.68\) / \(0.004\) ---
\textbf{collapsed (excluded)} \\
r2 & \(17.82\) / \(5.27\) --- healthy & \(9.83\) / \(1.08\) ---
healthy \\
\end{longtable}
}

VN is \(3/3\) healthy; plain is \(2/3\) healthy (r1 fails the filter by
latent collapse and is excluded \emph{before} the certificate is
evaluated, so the exclusion cannot select for or against the result).
The per-seed firm-up (calibrate on \(m\) sectors, test on the disjoint
rest, \(R{=}10\) sector selections, \(50\) episodes, four separation
modes; curve-level \(\Gamma_i\) score) is:

{\def\LTcaptype{none} % do not increment counter
\begin{longtable}[]{@{}
  >{\raggedright\arraybackslash}p{(\linewidth - 6\tabcolsep) * \real{0.2308}}
  >{\raggedright\arraybackslash}p{(\linewidth - 6\tabcolsep) * \real{0.2308}}
  >{\raggedleft\arraybackslash}p{(\linewidth - 6\tabcolsep) * \real{0.3077}}
  >{\raggedright\arraybackslash}p{(\linewidth - 6\tabcolsep) * \real{0.2308}}@{}}
\toprule\noalign{}
\begin{minipage}[b]{\linewidth}\raggedright
family/seed
\end{minipage} & \begin{minipage}[b]{\linewidth}\raggedright
\(m{=}1\): viol / \(\gamma_\alpha\) / ratio / \(u_{\ge.4}\)
\end{minipage} & \begin{minipage}[b]{\linewidth}\raggedleft
\(m^*_{0.4}\)
\end{minipage} & \begin{minipage}[b]{\linewidth}\raggedright
cost mode
\end{minipage} \\
\midrule\noalign{}
\endhead
\bottomrule\noalign{}
\endlastfoot
VN r0 & \(0\%\) / \(1.00\) / \(0.50\) / \(75\%\) & \(1\) & none \\
VN r1 & \(0\%\) / \(1.00\) / \(0.44\) / \(50\%\) & \(1\) & none \\
VN r2 & \(0\%\) / \(1.00\) / \(0.50\) / \(75\%\) & \(1\) & none \\
plain r0 & \(40\%\) / \(1.94\) / \(1.20\) / \(34\%\) & --- & violation +
slack (unsafe in \(m\le4\)) \\
plain r2 & \(0\%\) / \(1.00\) / \(0.38\) / \(33\%\) & --- & sharpness \\
\end{longtable}
}

All three VN seeds obtain a one-sector (\(m^*_{0.4}=1\)),
zero-violation, non-vacuous certificate; r1 is slightly less sharp
(ratio \(0.44\)), so \(\tau{=}0.4\) is the utility threshold at which
all three qualify. Neither healthy plain seed matches, with a
seed-dependent failure mode. Under the curve-level multiplier, plain r0
stays anti-conservative across the audited budget --- violation rate
\(40\%\to21\%\to11\%\) over \(m\in\{1,2,4\}\) (monotone but never
reaching \(0\), so \(m^*_{0.4}\) is undefined in range), despite a slack
\(\gamma_\alpha\approx1.9\); plain r2 is conservative but loses
sharpness (median ratio \(0.38<\tau\)). Seed-dependence is itself part
of the finding: non-equivariant orbit calibration is unstable, whereas
the equivariant certificate is consistent.

\textbf{Curve-level multipliers (and the retired proxy).} All 3D
conformal multipliers and violation rates above and in
Fig.~3 use the curve-level nonconformity score
\(\Gamma_i\) of Proposition B, computed by
\texttt{papers/figures/recompute\_gamma\_curve\_3d.py} from each
sector's \((\hat\delta,\hat\lambda)\) curve. The certified curve uses
the \(q90\) residual scale, which the per-sector artifacts do not store
directly; the recomputation therefore reconstructs it by solving for the
\(q90\) scale that \emph{exactly reproduces the stored certified
horizons}, and we verify that every recorded \(H^{\mathrm{cert}}\) (all
four \(\epsilon\) levels, all eight sectors, every checkpoint) is
recovered bit-for-bit before any \(\Gamma_i\) is computed. This is a
reconstruction for theory--implementation consistency, \textbf{not} a
re-fit of the 3D result: the audited horizons are inputs, not free
parameters. An earlier horizon-ratio diagnostic
(\(H^{\mathrm{cert}}/H^{\mathrm{meas}}\) with
\(H^{\mathrm{cert}}/\gamma\)) is kept only as an implementation sanity
check and is \textbf{not} used for any reported conformal claim. The
distinction is immaterial for fully conservative families, where every
\(\Gamma_i=1\) (so \(\gamma_\alpha=1.0\) for VN, plain r2, and the
entire 2D set), and matters only for plain r0 --- the one family with
raw anti-conservative cells (\(20/32\)). The residual plain-r0
violations under a positive \(\gamma_\alpha\) are the expected cost of
lacking zero-shot orbit transport (for a non-equivariant model the
wedge-calibrated sector is not exchangeable with the rest of the orbit),
not a failure of the conformal procedure. The horizon-division proxy was
optimistic here --- it reported plain r0 reaching safety by \(m{=}4\),
which the curve-level score does not.

\textbf{Scope.} This is a yaw (\(SO(2)\)) audit, not full \(SO(3)\)
tabletop dynamics, and it is a distributional audit of frozen models,
not a global reachability guarantee. It establishes that the
calibration-cost separation predicted by Corollary C \emph{does} appear
on a substrate where baselines are not already orbit-robust ---
complementing, not overturning, the 2D null.

\textbf{GPU/CPU equivalence.} The 3D audit uses a batched GPU rollout
for speed; we verified it computes the same certificate as the canonical
CPU \texttt{audit\_pair} on a shared checkpoint (vn\_r0) and the same
two episodes:

{\def\LTcaptype{none} % do not increment counter
\begin{longtable}[]{@{}lll@{}}
\toprule\noalign{}
quantity & CPU canonical & GPU batched \\
\midrule\noalign{}
\endhead
\bottomrule\noalign{}
\endlastfoot
\(\hat\delta\) & \(4.653\) & \(4.6526\) \\
\(\hat\lambda_1\) & \(0.0662\) & \(0.0662\) \\
horizon cells & \((2,3,1),(4,8,2),(8,8,3),(16,8,7)\) & identical \\
\end{longtable}
}

The \(\hat\delta\) difference (\(<10^{-3}\)) is floating-point reduction
order; the horizon cells the certificate actually emits are
bit-identical. Reproduce with
\texttt{experiments/p4\_3d\_calib\_cost.py},
\texttt{experiments/p4\_3d\_health.py},
\texttt{experiments/p4\_3d\_batch\_equiv.py}, and the analysis in
\texttt{papers/figures/analyze\_calib\_cost\_3d.py}.

\subsection{B. Proof of Theorem A (orbit
transport)}\label{b.-proof-of-theorem-a-orbit-transport}

\emph{Setup as in the main text:} \(G\) acts by
\(x\mapsto g\!\cdot\!x\), \(a\mapsto\sigma(g)a\), \(z\mapsto\rho(g)z\);
the transition \(T\), encoder \(E\), and predictor \(f\) are
equivariant, and \(\rho(g)\) is norm-preserving.

\emph{Proof.} Induction on \(t\). \textbf{Base case} \(t=0\): by
definition of the transformed initial condition, \(z_0^g=\rho(g)z_0\)
and \(\hat z_0^g=\rho(g)\hat z_0\). \textbf{Inductive step:} assume
\(z_t^g=\rho(g)z_t\) and \(\hat z_t^g=\rho(g)\hat z_t\). Using
equivariance of \(T\) and \(f\) and \(a_t^g=\sigma(g)a_t\),
\[z_{t+1}^g=T(z_t^g,a_t^g)=T(\rho(g)z_t,\sigma(g)a_t)=\rho(g)T(z_t,a_t)=\rho(g)z_{t+1},\]
\[\hat z_{t+1}^g=f(\hat z_t^g,a_t^g)=f(\rho(g)\hat z_t,\sigma(g)a_t)=\rho(g)f(\hat z_t,a_t)=\rho(g)\hat z_{t+1}.\]
By the norm-preserving metric,
\[\lVert\hat z_{t+1}^g-z_{t+1}^g\rVert=\lVert\rho(g)(\hat z_{t+1}-z_{t+1})\rVert=\lVert\hat z_{t+1}-z_{t+1}\rVert.\]
The per-step error is orbit-invariant for all \(t\le H\); since
\(\hat\delta\), \(\hat\lambda\), and the measured rollout error are
functions of these per-step errors, they are orbit-constant, hence so is
the raw curve
\(\widehat{\mathrm{Err}}(H)=\hat\delta\sum_{t<H}e^{\hat\lambda t}\).
Because \(\gamma_\alpha\) is a single scalar fixed on the calibration
set (not orbit-dependent), \(H^{\mathrm{meas}}\) and
\(H^{\gamma}=\max\{H:\gamma\widehat{\mathrm{Err}}(H)\le\epsilon\}\) for
every fixed \(\gamma\ge 1\) --- in particular \(H^{\mathrm{raw}}=H^{1}\)
and \(H^{\mathrm{conf}}=H^{\gamma_\alpha}\) --- are orbit-constant. Note
this is constancy of each horizon \emph{as a functional of the
orbit-constant curve}; we do not claim the non-identity
\(H^{\mathrm{conf}}=H^{\mathrm{raw}}/\gamma_\alpha\), which fails
because \(\widehat{\mathrm{Err}}\) is nonlinear in \(H\).
\(\blacksquare\)

\textbf{Exact vs.~approximate.} The theorem is exact under exact
equivariance. The implemented models are only approximately equivariant
(finite training; quantized fibers), so we measure the transport
residual
\(\bigl|\,\lVert\hat z^g_t-z^g_t\rVert-\lVert\hat z_t-z_t\rVert\,\bigr|\)
(the quantity Prop. D bounds) --- median \(1.1\%\), max \(4.1\%\) over
\(14\) audits --- quantifying how closely the implementation realizes
the theorem's assumptions. We distinguish this from the
\emph{out-of-wedge degradation} (orbit non-constancy of \(\hat\delta\)
\emph{without} transport), median \(2.2\%\) / max \(6.0\%\) over the
same audits: a related but looser diagnostic that does not isolate the
equivariance defect.

\begin{figure}
\centering
\includegraphics[width=0.66\linewidth,height=\textheight,keepaspectratio]{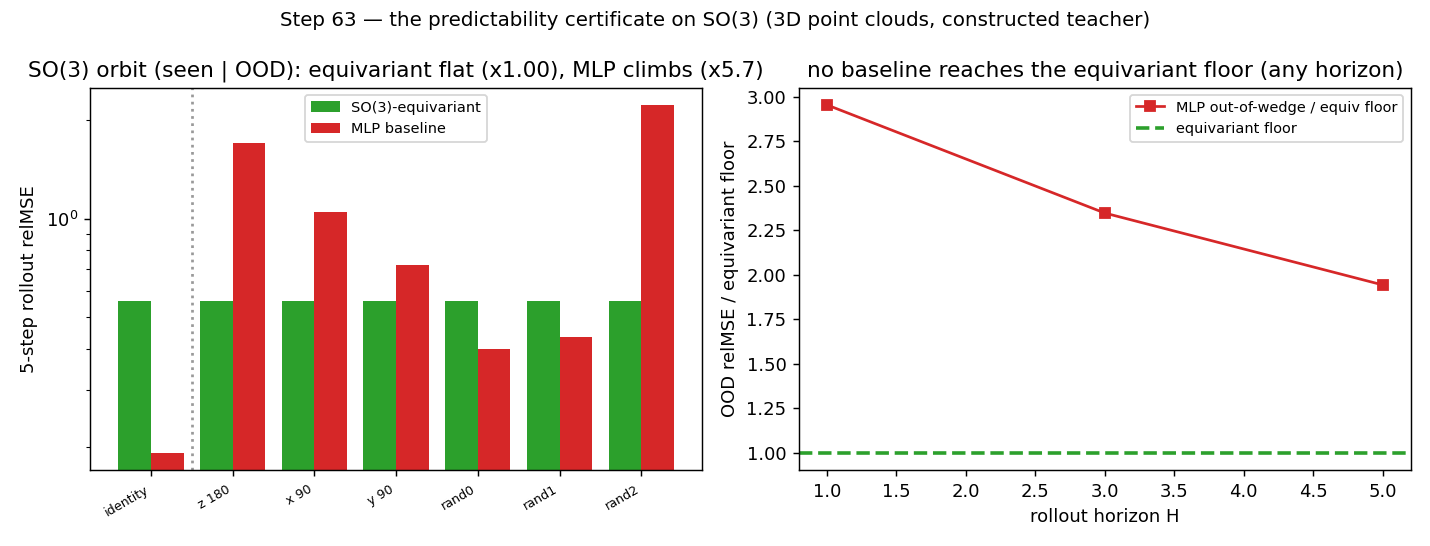}
\caption{Empirical orbit transport (the measured counterpart to Theorem
A, demoted from the main text). On a representation-level, synthetic
point-cloud \(SO(3)\) audit the equivariant model's rollout error stays
flat (\(\times 1.00\)) over the orbit while a non-equivariant baseline
climbs to \(\times 5.7\). This is representation-level \(SO(3)\),
\emph{not} full tabletop dynamics; dynamics-level claims are restricted
to the gravity-preserving yaw \(SO(2)\) subgroup (Section~5).}
\end{figure}

\subsection{C. Planning non-confirmation
diagnostics}\label{c.-planning-non-confirmation-diagnostics}

The 3D CEM-MPC non-confirmation (Section~6) rests on the well-powered
variance finding, not the correlation test. For completeness: across
\(n{=}10\) encoders at \(400\) episodes the per-seed optimal horizon
spreads over \(H\in\{2,3,4,8\}\) with \(3/10\) failing at every \(H\);
the formal trust-horizon / best-horizon association is Kendall
\(\tau_b=-0.15\) (permutation \(p=0.60\)), which is underpowered
(simulated power \(<0.4\) for \(\rho\le0.6\) at \(n{=}10\)) and
therefore reported as \emph{no evidence of a link}, not a demonstrated
absence. The cross-seed standard deviation stays in \(0.06\)--\(0.10\)
across \(150\to1000\) episodes --- data tames the bias, not the
variance. Reproduce with \texttt{papers/figures/hstar\_corr.py}.

\begin{figure}
\centering
\includegraphics[width=0.78\linewidth,height=\textheight,keepaspectratio]{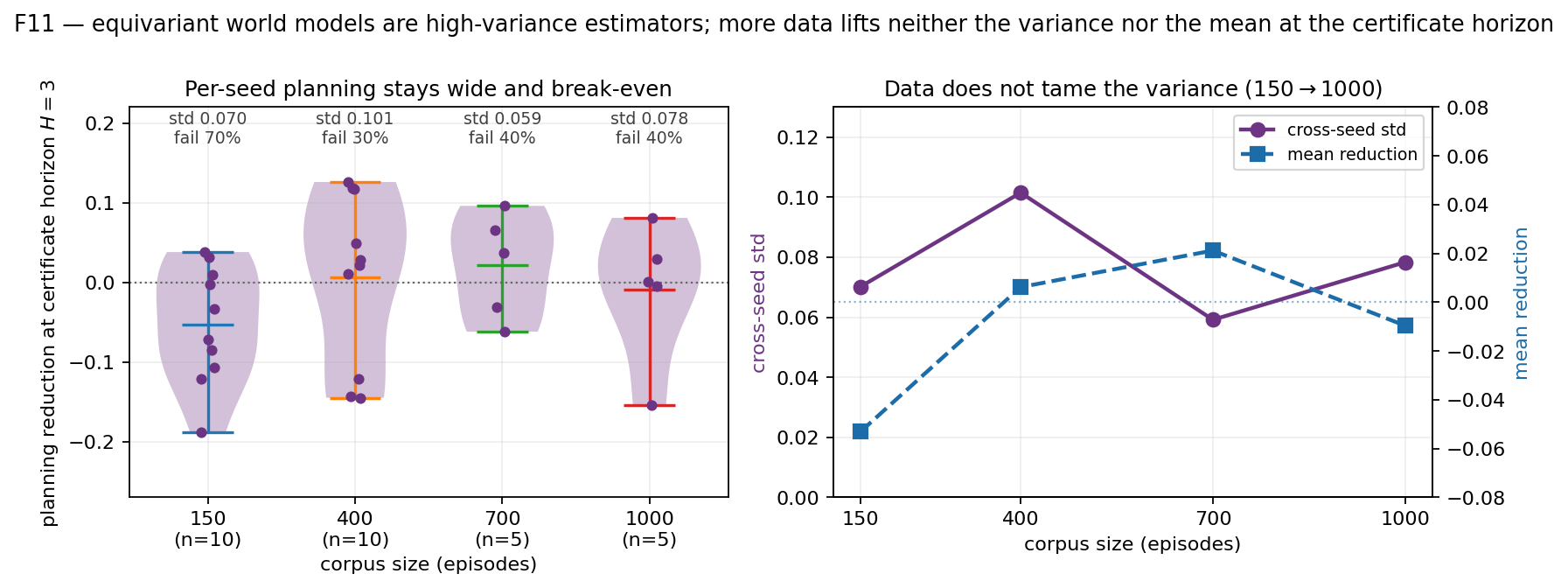}
\caption{Equivariance is a high-variance estimator --- the binding
downstream cost. Scaling the corpus \(150\to1000\) episodes (\(n{=}10\)
at 150/400) tames the bias (mean planning reduction moves toward
break-even, failure rate \(70\%\to\sim 40\%\)) but not the variance: the
cross-seed std stays in \(0.06\)--\(0.10\) at every data size.}
\end{figure}

\subsection{D. Structure--variance economics and the price of
pixels}\label{d.-structurevariance-economics-and-the-price-of-pixels}

These supporting characterizations motivate the Section~6 scope; they
are not the paper's headline and are demoted here from the main text.

\textbf{Structure--variance economics.} What equivariance buys, and
whether scale erases it, measured as curves rather than a single
benchmark point. (i) \emph{Prediction accuracy --- a low-data edge:} the
one-step ratio \(\hat\delta_{\mathrm{plain}}/\hat\delta_{\mathrm{eq}}\)
is \(2.4\)--\(3.3\times\) at \(\le 500\) episodes and \emph{closes} by
\(\sim 1000\) (a measured crossover; the scale objection of Brehmer et
al.~(2024) is correct on this axis). (ii) \emph{Readability and
certificate --- persistent:} the group coordinate is decodable only from
the equivariant latent (\(\theta\)-\(R^2\) \(0.91\) vs \(0.29\) at
\(40\)k; 3D TCP-\(R^2\) likewise), and does not collapse with data ---
exactly where the prediction edge does. (iii) \emph{The cost --- high
variance:} equivariant world models are high-variance estimators
(cross-seed \(\hat\delta\) spread; e2cnn-on-MPS non-reproducibility;
planner success swinging across seeds). The variance, not the bias, is
the binding constraint downstream --- the link to the Section~6 planning
non-confirmation.

\begin{figure}
\centering
\includegraphics[width=0.82\linewidth,height=\textheight,keepaspectratio]{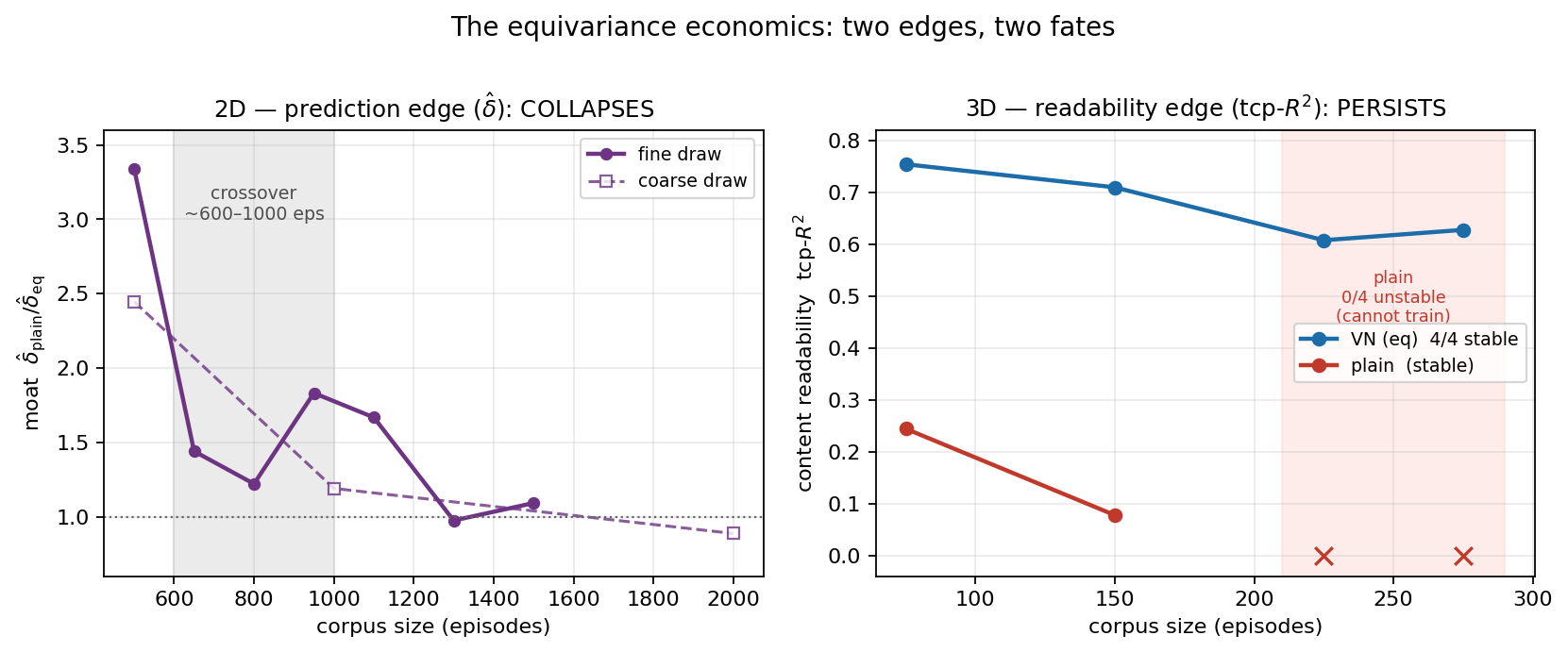}
\caption{Two edges, two fates. The 2D prediction edge
\(\hat\delta_{\mathrm{plain}}/\hat\delta_{\mathrm{eq}}\) collapses to
\(1\) past a \(\sim 600\)--1000-episode crossover (left; a low-data
phenomenon); the 3D readability edge (TCP \(R^2\)) persists ---
equivariant \(4/4\) stable, plain \(0/4\) (right).}
\end{figure}

\textbf{The price of pixels (under this substrate).} Neither lever
improved the latent's predictive precision: \(\hat\delta\) is flat
across corpus sizes \(c500\)--\(c2000\) on PushT pixels, and the
resolution lever \emph{backfires} --- the unit-safe within-rung ratio
\(\epsilon_{\mathrm{task}}/\hat\delta\) moves \emph{away} from the GO
regime at \(192\)px, where \(\hat\delta\) doubles (\(5.1\)--\(6.7\) vs
\(1.6\)--\(3.3\)) while \(\epsilon_{\mathrm{task}}\) stays flat (NO-GO
on all \(12\) cells; stage-A health passed \(4/4\), so the hit is on
predictive precision, not stability). Under this substrate, architecture
class, and sweep, the limiting factor is latent predictive precision,
not sensor resolution --- an empirical finding under the tested regime,
not a universal claim about pixel-level modeling.

\end{document}